*Research Article*

# A Framework for Analyzing Fog-Cloud Computing Cooperation Applied to Information Processing of UAVs

**Milena F. Pinto**,[1] **André L. M. Marcato**,[1] **Aurélio G. Melo**,[1] **Leonardo M. Honório**,[1] **and Cristina Urdiales**[2]

[1]*Department of Electrical Engineering, Federal University of Juiz de Fora (UFJF), Juiz de Fora-MG/36036-900, Brazil*
[2]*Department of Electronics Technology, University of Málaga, Málaga 29071, Spain*

Correspondence should be addressed to Milena F. Pinto; milena.faria@engenharia.ufjf.br

Unmanned aerial vehicles (UAVs) are a relatively new technology. Their application can often involve complex and unseen problems. For instance, they can work in a cooperative-based environment under the supervision of a ground station to speed up critical decision-making processes. However, the amount of information exchanged among the aircraft and ground station is limited by high distances, low bandwidth size, restricted processing capability, and energy constraints. These drawbacks restrain large-scale operations such as large area inspections. New distributed state-of-the-art processing architectures, such as fog computing, can improve latency, scalability, and efficiency to meet time constraints via data acquisition, processing, and storage at different levels. Under these amendments, this research work proposes a mathematical model to analyze distribution-based UAVs topologies and a fog-cloud computing framework for large-scale mission and search operations. The tests have successfully predicted latency and other operational constraints, allowing the analysis of fog-computing advantages over traditional cloud-computing architectures.

## 1. Introduction

In recent years, UAVs have been used in various applications such as monitoring [1], surveillance [2], topography [3], and archaeological exploration [4]. This versatility is explained by the ability of UAVs to perform complex activities with maneuvering flexibility and low-cost flight. In regular missions, the operator controls the vehicle position in every situation. However, when the mission is semiautonomous, and the operator is responsible for just a few tasks, such as taking off and landing the aircraft while this one carries out the flight autonomously through waypoints. In fully autonomous reactive missions, the trajectory is created onboard the aircraft and the mission is performed without the operator nearby. In this sense, the autonomous aerial robotics are connected to a supervision system (i.e., ground station—GS) that is usually located at the cloud and is responsible for all high-level processing.

However, this cloud-based approach may be inappropriate for sensitive real-time systems once the exchanged amount of data among the devices would generate higher costs of communication bandwidth, lack of mobility, communication delay, energy constraints of embedded systems, and information redundancy [5]. To mitigate these issues, a new trend of computing paradigm is to make the computation and storage close to the end-devices, which in this particular case are the drones. Fog computing arises as an intermediate layer between cloud and end-devices to improve latency, power consumption, scalability, and efficiency. This technology allows overcoming the limitations of centralized cloud computation by enabling data acquisition, processing, and storage at fog devices [6, 7].

Under these assumptions, this research work proposes a fog-based framework focused on cooperative-based UAVs topologies. This approach uses a UAV as a fog computing node to provide services. The services should be deployed in this node along with a filtering and clustering mechanisms. The proposed filtering methodology is an importance-based classifier that allows critical information to be delivered in accordance with application requirements. The fog computing can also be a supervision and coordination mechanism, which is not presented in other architectures [8–15].

To validate this methodology, a mathematical model capable of simulating distributed systems is proposed. This model captures the behavior of main feasibility parameters such as latency and throughput in the fog-cloud computing proposed architecture. Therefore, it is now possible to evaluate different framework designs and choose the best one for each solution. As a result, it is possible to compare the proposed fog-based approach with the traditional cloud-based ones.

*1.1. Motivation.* Some applications are extremely susceptible to delays. For example, Search and Rescue (SAR) and Inspection are generally executed at remote locations with low communication resources. In this context, most of the decisions are target detection [16, 17] and team coordination [18]. These tasks are particularly delay-sensitive. For instance, in target detection, the object can be lost in fractions of seconds in case of improper detection.

The research presented in [12] suggests that one second of delay is already a challenge in cloud-supported applications and values lower than 100 milliseconds are unattainable. This issue impacts real-time applications and reduces the ability to control systems. Some applications with UAVs are susceptible to this kind of problem, which degrades substantially the quality of the missions subjected to delays. Besides these limitations, many of these applications require a large amount of data, especially for streaming videos to GS for image processing or monitoring [19]. The operation at places with low communication infrastructure can also bring bandwidth challenges that have to be properly addressed.

Another important motivation is the power consumption in fog and cloud. The mission time and services provided by the aircraft are primarily limited by the amount of available energy. For instance, quadrotors typically have flight times lower than 25 minutes. In this sense, any optimization can greatly improve the system overall performance.

The motivation for this work's development comes from the nonusual fog-cloud computing applicability with multiple UAVs. The main challenge is to propose an architecture to evaluate the applicability of fog-cloud computing cooperation aiming at optimizing latency while keeping throughput and power consumption under a range. Based on the mentioned problems, this work highlights the importance of addressing three research challenges:

(1) Data processing closer to end-devices.
(2) A platform to support fog-cloud computing deployment with minimum power consumption and throughput usage.
(3) A model to analyze the efficiency of fog-cloud computing collaboration and its requirements for UAVs.

*1.2. Work Contribution.* The UAV assigned as the head coordinator is in charge of the fog-cloud computational offloading. The head coordinator analyzes the data to process or to transform them into chunks of selected information before transmitting to the cloud. This work considers that each UAV has an embedded framework responsible for controlling the task planning, mission, and flight parameters to ensure autonomous operation. Besides, this embedded framework contains a fog device managing the information of its respective group. Considering this scenario, this research work's contribution can be summarized as follows.

(i) A new framework layout to overcome the throughput and latency limitations involving multiple aircraft during missions in areas with restricted communication infrastructure.

(ii) A model to analyze the feasibility of fog-cloud computing cooperation in UAV context taking into consideration latency, energy, and throughput constraints.

(iii) This research has conducted simulations to validate the model, which shows that the transmission and processing delays, energy consumption, and throughput can be reduced, saved, and optimized, respectively.

*1.3. Organization.* The remainder of the paper is organized as follows. The background and related works are presented in Section 2. Section 3 discusses the advantages and the constraints that can be optimized when applying fog computing to work cooperatively with cloud data centers. Section 4 details the problem formulation of this work decomposing the system model in three subproblems: latency, throughput, and power constraints. The simulations and discussing of the numerical results are presented in Section 5 and Section 6. The concluding remarks are conducted in Section 7.

## 2. Background and Related Works

The cooperation among multi-UAVs to perform tasks has several advantages. For instance, the quality and the time to complete a mission can be improved by mutual cooperation. Also, the parallel execution of tasks can increase the probability of mission completion [29]. Besides, flexible platforms can be reallocated UAVs when damage occurs to ensure mission completeness. A drawback of cooperative-based task performing is the complexity of sharing resources and information. For instance, each UAV perform minor decisions (e.g., battery information, and collision avoidance) and central intelligence deals with mission strategies such as coordination, supervision, and high-level information analysis and decision-making. In this scenario, a full communication among agents is required. However, in some cases, the challenges to the cloud approach have higher costs of communication bandwidth, communication delay, embedded systems energy constraints, and information redundancy could be inappropriate.

Current architectures for the development of autonomous UAVs [30] do not incorporate technology to overcome the limitations of centralized cloud computation. For instance, the works [31, 32] show the possibility of cloud computing failure due to network impairments. Fog-cloud cooperation arises as an opportunity to improve latency, power consumption, scalability, and efficiency for information exchange of end-devices [14]. To measure the effectiveness of an approach involving embedded systems, it is necessary to have a mathematical model to validate the distributed architecture. To the best of the authors' knowledge, there is no current proposed architecture to develop UAV cognitive systems that incorporate fog concepts in itself as part of the processing stack [30, 33, 34] neither a specific mathematical model to validate all necessary requirements in this context [8–15].

*2.1. Fog Computing.* Several pieces of researches have been published to formally define the fog computing with its respective challenges. Its benefits and issues are surveyed in Dastjerdi and Buyya [8] and Mouradian et al. [35] by presenting an overview of this topic along with its characteristics. Some discussions of challenges, application scenarios, and emerging trends can be found in [36–38], respectively. However, there is a difficulty in using the available fog platforms in remote areas due to an unreliable connection.

A practical application of fog computing requires an architecture to achieve the proposed goals. For instance, the work [39] presents a fog architecture with a flexible software-defined network (SDN) to programmatically control networks. The devices in this network should present a flexible self-organized structure to allow the insertion of fog nodes. The work [40] introduced the concept of the virtualization of services where a fog implementation based on ROS performs the services for a network of robots. This implementation resembles the one proposed in this research work.

Few other topics related to the fog application in this context are worth mentioning. An implementation to minimize the services delay is presented in [41]. This work proposes a policy for fog nodes considering queue length and different request types with variant processing times. Other active topics are the complex requirements to obtain a highly reconfigurable network [42] and the paradigm of implementing shared services and resources [43, 44]. These subjects are essential to the UAV-Fog operation but are not discussed in the present work.

*2.2. Latency, Throughput, and Power Constraints.* As explained, the above architecture aims at optimizing three sub-problems. Thus, instead of comparing the proposed model with others, this section compares each desired parameter. The first and most important one is the latency perceived by the end-user. Currently, in several applications, the interaction between a group of UAVs and cloud is performed individually. However, this may be inefficient and costly if the number of data increases, which may also present high redundancy.

The latency is modelled in slightly different ways in the literature. In [10], the latency is calculated by combining the time required for transfer data between nodes, the processing time, and the period related to balance the services among nodes. In [45], these previous factors are considered along with a nonlinear component that accounts for differences in transmission channels such as the queuing order along nodes. Other models may also consider factors like average transmission error in the networks, inter-UAV communication latency, and the time required for clustering data in fog [13, 46]. There are also different modelling techniques which include statistical analysis and modelling based on queue theory where the latency is calculated as the average response time in a queue model M/M/1 [11].

The second considered characteristic is the throughput between fog and cloud computing. The throughput may be affected by some factors, including limitations in the hardware, available power processing, and end-user behavior. Despite being a challenge to fog application, these factors are usually not modelled in many works and can also be compressed into a single rate of error in the transmission channel [21]. In [22, 47], the throughput of 3G and 4G networks is analyzed as a network peak data ratio distributed among users.

The last analyzed characteristic is power consumption. The work [9] presents a model to analyze the power consumption and to evaluate the tradeoff between power usage at fog and cloud nodes during the network operation. The model considers the fog computing as a data center. This consideration provides preloaded content to end-users. Despite the good results, the model looks at the problem to save energy at the cloud servers. In this proposed paper, the problem restriction is to optimize power consumption at the fog devices to extend the UAV flight time. The power consumption of data-forwarding is analyzed in [10, 11]. The fog device model is seen as a resource with unlimited access to power supply, which is not ideal for an embedded application. In this sense, the current state of the art can be improved with models that reduce latency while still maintaining an optimal power consumption in the fog level.

## 3. General Framework

This section discusses the proposed model, addressing all its components. The problem representation is illustrated in Figure 1. Each UAV is responsible for determining its trajectory autonomously based on the assigned task. Besides, each UAV captures images and information required for its missions. During the task execution, some commands are the responsibility of the GS, for example, supervision, position monitoring, data video storage, and task assignment. These basic definitions lay the groundwork for defining the architectural problem, i.e., defining how components are organized. The next topics discuss the main aspects related to the proposed problem.

*3.1. Fog Node Localization.* The localization of the fog computing node is a key component in the architecture definition. There are two considered possibilities. First one is the placement of the fog node on the ground, geographically near the UAV group. The second possibility is to embed the fog device in the UAV coordinator that will move together with its team during the mission execution. The main benefit of the first option is to avoid power restriction due to the possibility of connecting the fog device to power stations or generators.

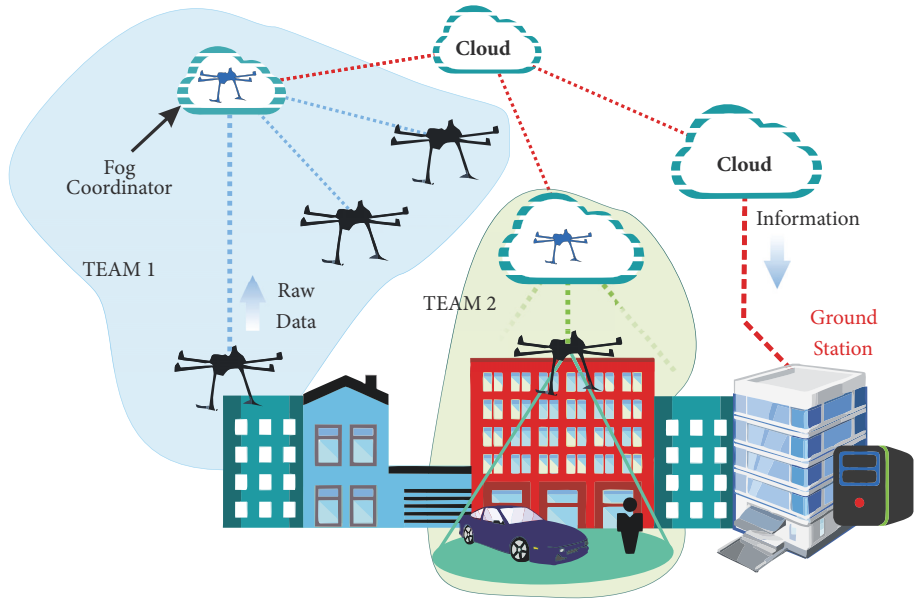

Figure 1: Problem representation.

As a tradeoff, this configuration creates data transmission restrictions once the objects on the ground can interfere with the network signal.

The deployment of a fog node makes the UAV architecture flexible and more autonomous. An example is the flight time of the head coordinator that does not affect the mission due to the fact that the all aircraft executing the missions share the same power restrictions.

### 3.2. Fog Node Services.
Initially, the first implementations of fog computing were responsible only for information clustering. However, the current applications pack several types of preprogrammed services capable of handling the incoming information directly at the fog device. This work uses the second approach. The head coordinator will manage the information of its group to provide services and to cluster the information into the cloud.

It is important to know the requirements of UAVs to determine which services are necessary at the fog level. In this work, the aircraft can operate autonomously; i.e., they can perform flight control and data gathering and take decisions regarding their tasks. Especially for SAR context, the UAVs need to flight along a certain path to capture images for objects recognition. Besides, they need to decide whether the mission is accomplished or not based on the acquired information. An architecture capable of supporting this level of automation was proposed by the authors in [2].

In many architectures, the GS assists the aircraft when activities require a certain level of high cognition. Usually, the supervisory system performs task planning, monitoring, path planning, among others. However, most of these activities do not require direct human intervention and can also be performed by another autonomous system. In the approach of this work, some services are deployed on the fog computing node whenever is possible to reduce the fog-cloud throughput (e.g., data storage). Other tasks such as the mission goals definition and supervision are processed in the cloud.

The proposed task distribution requires that important information is shared with the cloud and a great part of the data stays at the fog level. In this sense, an algorithm must classify the incoming data to determine which ones should be sent to the cloud. Based on the premises that each UAV can take decisions related to the task execution, it is possible to say that critical mission data can also be flag accordingly to its importance.

This explanation is represented in Figure 1. A coordinator located at the fog level manages the information of the nearest nodes. The data that is not processed and stored locally in the coordinator is forwarded to the supervisory system located at the cloud for further processing.

### 3.3. Fog Node Characteristics.
The hardware characteristics are important to conduct the analysis of the proposed architecture. Initially, the methods for data transmission regarding the inter-UAV and fog-cloud communication are selected. Each technology should provide different data rates and transmission ranges. Table 1 shows a list of key characteristics for the typical wireless communication technologies. The selection of proper hardware for UAV application should consider the relation between energy/coverage for the large flight times.

The services specified in the last topic can be deployed to the nodes using different methods ranging from virtualization of components (i.e. containing apps and libraries) to direct programming of the services in the host operating system. The selected system-on-a-chip (SoC) should be power efficient and support the fog services and the employed methods. Several components can be applied to this task and their common characteristics are presented in Table 2.

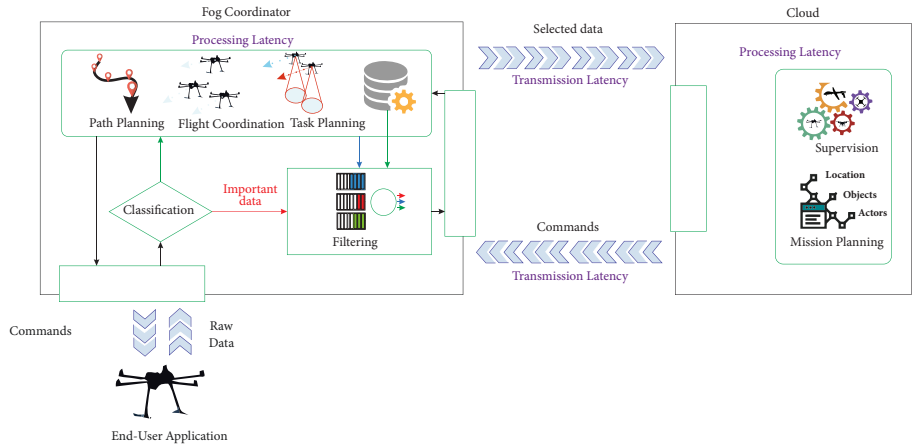

Figure 2: Fog-cloud computing network dataflow.

Table 1: Characteristics of typical wireless standards [20–24].

|  | Bluetooth | Wi-Fi | HSPA | ZigBee |
| --- | --- | --- | --- | --- |
| Coverage | 100 m | 0.1 – 2 Km | 5Km | 1.2 – 14 Km |
| Throughput | 22 Mbps | Up to 300 Mbps | 5.76 – 11 Mbps | 0.25 – 72 Mbps |
| Frequency | LF 120-134Khz HF 13.56MHz UHF 850-960MHz | 2.4, 5GHz | TDD 1.85-3.8GHz FDD 0.7 – 2.6GHz | 0.9, 1.2, 2.4 GHz |
| Energy Efficiency | High | Low | Depends on the Signal Strength | Depends on the Model |

Table 2: Common characteristics of SoC [25, 26].

|  | ARM | x86/x64 + GPU | FPGA |
| --- | --- | --- | --- |
| General Task Efficiency | Medium | High | Low |
| OpenCV Efficiency | Low | High | Medium |
| Energy Efficiency | Medium | Low | High |
| Implementation Complexity | Low | Low | High |

*3.4. Fog Computing Architecture.* This approach is detailed in the architecture presented in Figure 2. The idea is based on [2, 42, 44]. Some of the data is sent to the supervision devices such as images and the geographical positions. However, all data is marked according to the type of message. The data coming from the different aircraft is received in the coordinator and classified based on their importance. Posteriorly, they are sent either to local processing or to filtering. If classified as important, the data goes directly to the filtering block to be clustered and sent directly to the GS. Otherwise, the data is transferred to local processing, which can process or store. This means that the coordinator has similar algorithms to the ones employed on the GS, i.e. the fog level can assist the UAVs during the execution of their tasks. Image and other data can also be stored at this level to be recovered by the GS in the future.

Figure 2 also shows the latencies considered in the model. The execution of the tasks available in the head coordinator along with the data classification and filtering will result in the processing latency at the fog level. The transmission latency is a function of the channel throughput and its characteristics. The processing latency in the cloud is a result of the services execution time provided by the cloud data centers.

The comprehension of the dataflow can be improved by analyzing two SAR examples. The first one is about non-important data during the mission. In this example, the UAVs should regularly update their positions to enable continuously tracking by the GS. As this regularly updating can account for a large amount of data over time, the local processing can analyze if the position has changed enough to justify this action. Another example is the fog device taking actions locally and informing the GS in case that the aircraft is lost or did not update the position regularly. In SAR and Inspection missions, the recognition of a target is an important data that should be informed to the GS immediately.

## 4. Feasibility Modelling

The framework described in the last section presents an alternative of including the fog in the UAV decision structure. However, it is important to analyze the feasibility of its implementation. This section investigates these aspects and presents a discussion of fog possible benefits.

*4.1. Constraints.* Few considerations are required to model the system. First, the physical constraints of the fog device limit the processing. Thus, the workload $l_i$ is allocated by the processing capability of the fog $v_i^{fog}$. Other limitations are the fog-cloud communication stability and the average throughput required by the fog device that should be lower than network throughput.

*4.2. System Model*

*(1) Throughput.* For a fog-cloud computing collaboration, the throughput $B_i^{fog}$ can be modelled as a function of the average arrival rate $\delta_i^{fog}$, the average packet size $s_i^{fog}$, and the rate of packets rejected for fog processing $(1 - r_i)$, such as in [10]. Note that the physical limitations of the system may affect the throughput of the communication, such as the available processing power and the end-device behavior.

$$B_i^{fog} = \delta_i^{fog} \cdot (1 - r_i) \cdot s_i^{fog} \quad (1)$$

Note that the intention of minimizing this equation is to control the amount of data forwarded to the cloud. This can be more restrictive if only mobile cellphone networks are available or less restrictive if more modern and energy efficient infrastructure is available

*(2) Power Consumption.* Several factors may influence the power consumption in fog and cloud. These factors include the used algorithms and the environment temperature. However, the main factor is the computational load due to the information coming from the end-devices. For simplification, the proposed model considers the rate of data accepted for processing (i.e. Bits/second) and uses a standard value for the idle power consumption as in [9]. This model is shown in Figure 3.

Then, the power consumption of a fog device $i$ is modelled as a linear function of the data accepted for processing rate $r_i$ in respect with the predetermined processing efficiency $\gamma > 0$ and the idle power usage $\theta > 0$. The consumption boundary is mainly determined by the processor Thermal Design Power (TDP) or $E_i^{fog} \leq TDP$.

$$E_i^{fog} = \gamma \cdot \delta_i^{fog} \cdot s_i^{fog} \cdot (r_i) + \theta \quad (2)$$

Note that the power usage at the cloud is not considered. This is to ensure a power consumption optimization at the fog level, which will result in increasing flight time for the aircraft. This equation captures the amount of information selected for local processing. Since the most relevant information is desired to only be processed locally, thus, minimizing the energy consumption will penalize the local data processing to ensure this condition.

*(3) Latency Computation of Fog-Cloud Computing.* The latency $\omega_i^{fog}$ in the fog device should be lower when few packets

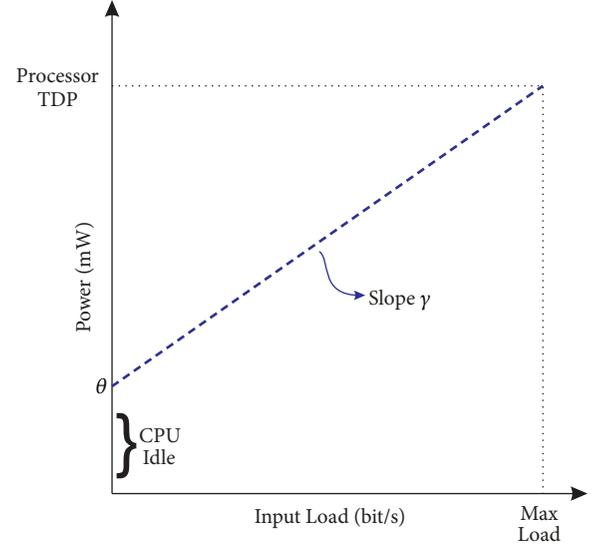

FIGURE 3: Power consumption at the fog device.

are transferred to the cloud and should grow exponentially if the number of packets exceeds the processing capability $v_i^{fog}$. The process should be stable if the average arrival rate $\delta_i^{fog}$ at the fog level is lower than its processing capability.

$$\omega_i^{fog} = 2^{(\delta_i^{fog} \cdot r_i)/v_i^{fog}} \quad (3)$$

If the average is selected to contain only stable arrival rates ($\delta_i^{fog} < v_i^{fog}$), the function can be approximated as

$$\omega_i^{FOG} \approx \frac{\delta_i^{fog} \cdot r_i}{v_i^{fog}} \quad (4)$$

The latency at the cloud $\omega_i^{cloud}$ is represented by the relationship between the data transmitted to it, the network throughput, and the processing capability of the cloud data center. The amount of data sent to the cloud can be determined by the following function $s_i^{fog} \cdot \delta_i^{fog} \cdot (1 - r_i) = d_i^{cloud}$ and the amount of data transmitted between fog and cloud devices by $t_i^{fog-cloud}$. After being processed at the cloud, a fraction of the input data $a_i^{cloud}$ is sent back to the fog. The perceived latency computed from the cloud is detailed in

$$\omega_i^{cloud} \approx \frac{d_i^{cloud}}{2 \cdot t_i^{fog-cloud}} + \frac{d_i^{cloud} \cdot a_i^{cloud}}{2 \cdot t_i^{fog-cloud}} + \frac{d_i^{cloud}}{2 \cdot v_i^{cloud}} \quad (5)$$

where $v_i^{cloud}$ represents the processing capability of the cloud. The total latency is given by (6), where $\omega^{fog}$ and $\omega^{cloud}$ denote the respective latency in the fog and the cloud.

$$D_i^{AVG} = AVG\left(\omega^{fog}, \omega_i^{cloud}\right) \quad (6)$$

Two state parameters are defined to simplify the problem. The first one $x_1 = \delta_i^{fog} \cdot r_i$ represents the total quantity of packages in the fog device and the second parameter $x_2 = \delta_i^{fog} \cdot (1 - r_i)$ is the total of packages in the cloud. This problem can be rewritten to minimize three objective functions, as presented in

$$\min \left[ B_i^{fog}, E_i^{fog}, D_i^{AVG} \right]^T$$

$$= \begin{bmatrix} 0 & s_i^{fog} \\ \gamma & 0 \\ \dfrac{1}{2 \cdot v_i^{fog}} & \dfrac{t_i^{fog-cloud} + v_i^{cloud}}{2 \cdot t_i^{fog-cloud} \cdot v_i^{cloud}} \end{bmatrix} \begin{bmatrix} x_1 \\ x_2 \end{bmatrix} + \begin{bmatrix} 0 \\ \theta \\ 0 \end{bmatrix} \quad (7)$$

As can be seen, the model evaluates the tradeoff among bandwidth, energy in the fog node and the delay perceived by the UAV. This model should capture the variables behavior for a given set of parameters from the application when using the proposed framework. Moreover, the goal of reducing latency forces an amount of data to be accepted for local processing in the fog computing node. As this variable grows, the energy constraints should keep this balanced within an acceptable boundary due to local processing capability and energy consumption.

*4.3. Modification 1.* The previous model represents one of the possibilities for analyzing the fog-cloud computing viability. However, other few considerations can be also relevant depending on the hardware requirement. One important assumption is to consider the power consumption almost linear with respect to data processed in the fog device. However, this may not be true in every case. The power consumption required to transmit data to the cloud can also be a relevant part of the total power used in the head coordinator. Thus, a modification using an extra parameter $\rho$ to represent the power required for data transmission is shown in

$$E_i^{fog} = \gamma \cdot \delta_i^{fog} \cdot s_i^{fog} \cdot (r_i) + \rho \cdot \delta_i^{fog} \cdot s_i^{fog} \cdot (1 - r_i) + \theta \quad (8)$$

*4.4. Modification 2.* The deterministic comportment of the transmission latency is an important feature for analyzing the model behavior in the experiments. However, in some situations, it may be useful to analyze the latency behavior in a more realistic state. The modification of the model includes a random latency component $f(d_i^{cloud}, t_i^{fog-cloud})$ with a mean value equal to 1 and limited variance. This variation is presented in

$$\omega_i^{cloud} \approx \dfrac{d_i^{cloud}}{2 \cdot t_i^{fog-cloud}} \cdot f\left(d_i^{cloud}, t_i^{fog-cloud}\right) + \dfrac{d_i^{cloud} \cdot a_i^{cloud}}{2 \cdot t_i^{cloud-fog}} + \dfrac{d_i^{cloud}}{2 \cdot v_i^{Cloud}} \quad (9)$$

## 5. Experimental Design

An environment was deployed on the software MATLAB version R2016a [48] to simulate the proposed model. Besides,

Table 3: Bitrate required for common image sizes used in UAV applications [27, 28].

| Resolution | Minimum bitrate | Maximum bitrate |
|---|---|---|
| 360p | 400 Kbps | 1.000 Kbps |
| 480p | 500 Kbps | 2.000 Kbps |
| 720p | 1.500 Kbps | 4.000 Kbps |
| 1080p | 3.000 Kbps | 6.000 Kbps |

Table 4: Network characteristics of common mobile standards.

| Network | Latency | Throughput (Uplink) |
|---|---|---|
| GSM | 600-750 ms | 40 Kbps |
| UTMS | 500-750 ms | 384 Kbps |
| HSPA | 150-400 ms | 5.76 Mbps |
| HSPA+ | 100-200 ms | 11.5 Mbps |

a multiobjective optimization function was applied in the algorithm using a Genetic Algorithm [48]. In this experimentation, a broad range of common requirements is selected for the aircraft application and network infrastructure. Moreover, few assumptions about the data are required due to the dependency of a scenario to apply the model. Then, the experimentation consists in selecting a given configuration (i.e. packet processing capability at fog, network throughput and arrival rate) and after that, the algorithm works finding solutions for a set of different work distributions between fog and cloud devices.

*5.1. Environment Assumptions.* Missions using UAVs require different types of data. Image is one of the most demanding ones. Thus, it is necessary a bandwidth with a capacity to transmit images with different qualities during the tasks. The bandwidth can be used as a parameter to analyze the throughput requirement for the system operation. In this work, the simulations use bandwidth related to video transmission varying from a single aircraft in 360p up to a team of 6 UAVs transmitting video in 1080p [27, 28] as can be seen in Table 3.

The throughput and inherent latency of the network depend mostly on the network protocol. In this sense, the throughput and latency are studied selecting different types of networks ranging from Edge to 4G, as shown in Table 4 [21, 22].

The fog processing capability depends on both architecture and processor. It is important to determine the maximum of load that the fog computing can absorb before becomes unstable or present a high processing latency. In this way, the utilization of fog computing in situations where the system cannot process enough packets may degrade the average latency. In this case, the application turns impracticable. Thus, in this work, the processing capability changes from 25% to 100% of incoming packets.

The efficiency of the processor is the most difficult parameter to determine due to its dependence on many factors, such as thermal processor efficiency, operational system and the services provided by the UAV coordinator.

This work considers the processor parameters matching an x86 processor model z3775g when changing from idle to full utilization. The power consumption was measured using the operational system Ubuntu 14 with LXDE. During the measurements, the system performed image processing, data storage, and general data processing.

There are many requirements for the architecture implementation. For example, the UAVs intercommunication is a critical issue. However, it will not be a subject of this work. In this paper, this communication is considered as a high-speed local connection with low power [42] and should ensure that this data exchange has a minimal delay and it does not significantly affect the proposed problem. The information offloaded between fog and cloud devices are provided by a different interface that uses mobile telecommunication, e.g., GPRS networks.

## 6. Results and Discussions

This section presents the results and the respective analyses of the developed model efficacy. Some numerical assumptions about the problem were presented in the previous section to show how the model works without defining a specific scenario for the problem. The first parameter is the different levels of data traffic generated by a given number of UAVs. The second one is the characteristic of some mobile network standards. Lastly, the third parameter simulates the levels of processing capability at the fog level.

These parameters illustrate the behavior of the objective functions. For example, we can analyze data traffic/throughput rate versus fog-cloud computing viability. A question to ask is how much rate among data traffic and throughput must increase before fog computing becomes viable? These analyses should assist the decisions boundaries understanding for each objective function presented in (2), (3), and (7).

The first result is in Figure 4. The red and blue arrows help in visualizing the increase direction of fog processing rate and throughput, respectively. The parameters were assigned with colors that represent their original configurations. In this figure, the throughput and processing rate are represented respectively by blue and red colors. Note that the values of throughput and processing rate are constant for a set of points with the same color. The throughput affects the latency baseline, as indicated by the blue arrow. The latency behavior in respect with the workload distribution depends on the fog processing rate; i.e., if the fog processing is at a certain minimum value, the latency will decrease as the workload is added to the fog level. Figure 5 shows the energy efficiency and throughput variation. Note that both parameters present a simple exchange. This is related to the problem design that considers only the throughput between fog and cloud computing and power consumption at the fog computing.

The variation of the transmission rate between fog and cloud computing is shown in Figure 6. This variation analyzes the system behavior as the performance of the communication structure improves. The lines in Figure 6 turn around a point with 100% of the workload in the fog device. However, in Figure 4, the curves turn around 100% of workload in the cloud. This indicates that the transmission rate behaves concurrently with the fog processing rate presented in Figure 5. The latency at the fog will be improved as more work is transferred to the cloud.

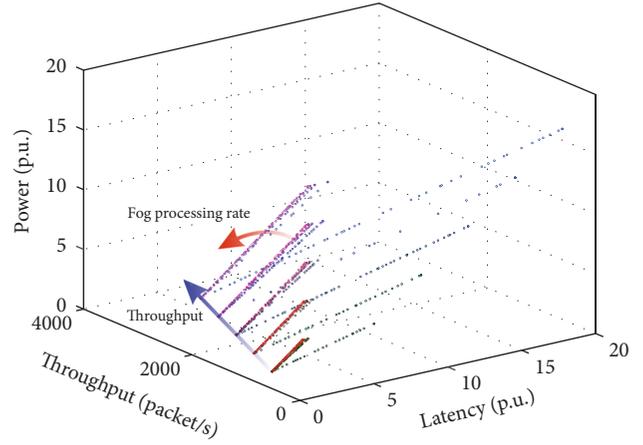

FIGURE 4: Optimization for packets arrival rate at fog device.

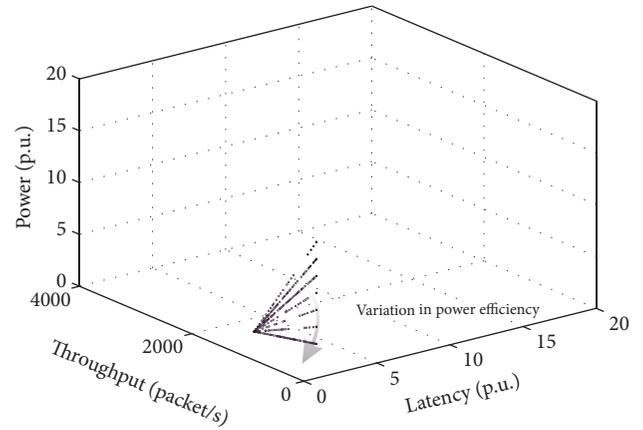

FIGURE 5: Energy efficiency behavior.

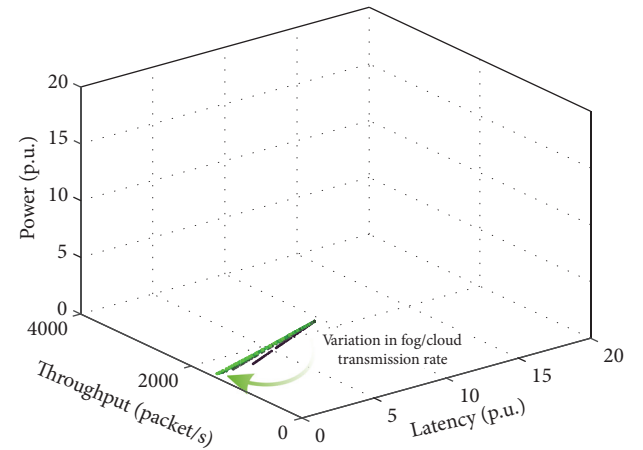

FIGURE 6: Optimization of fog and cloud transmission rate.

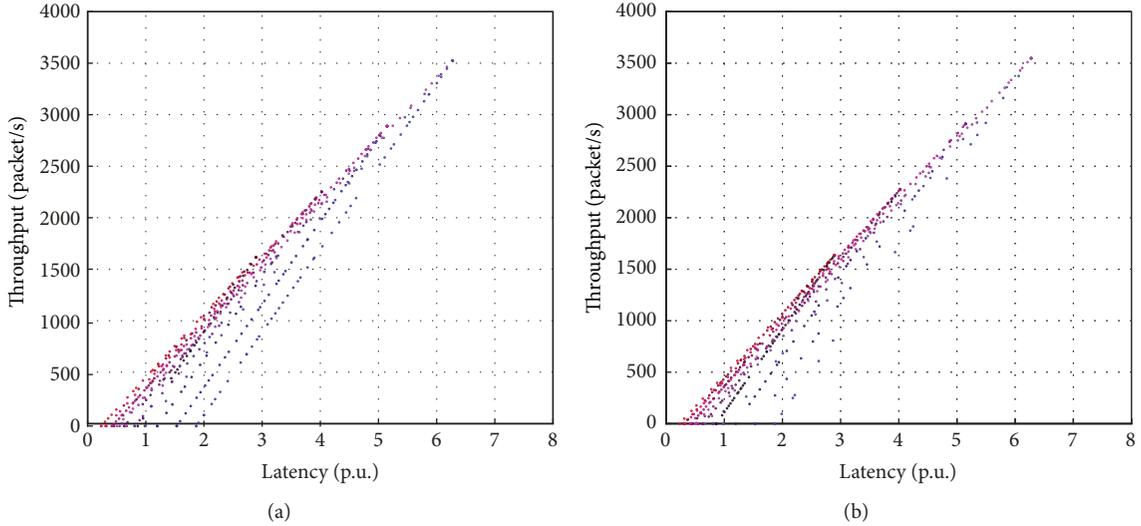

Figure 7: Throughput and latency for a variable number of incoming packets. (a) Original model. (b) Modification 1.

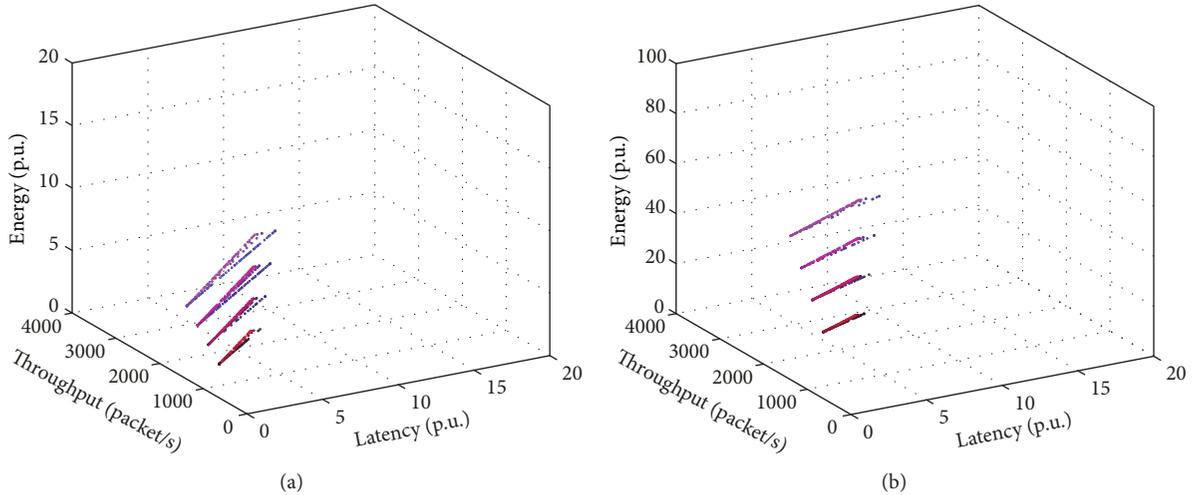

Figure 8: Energy, throughput and latency for a variable number of incoming packets and fog processing rate. (a) Original model. (b) Modification 1.

*6.1. Results for Modification 1.* As mentioned before, the first modification in the proposed model considers the power required to offload data from fog to the cloud. The throughput and latency behaviors for a variable number of incoming packets are shown in Figure 7(a). Figure 8(a) exhibits the energy, throughput and latency for a different number of incoming packets and fog processing rate. For comparison purposes, Figures 7(b) and 8(b) exhibit the model behavior without considering the power required for data transmission. The results of Figure 7 do not show significant changes. This means that the relationship between latency and throughput is not affected by this parameter. However, in Figure 8, this parameter changes the minimum value of energy consumption for a determined configuration.

*6.2. Results for Modification 2.* The second modification of the model illustrates the effect of randomness in the network on the system operation. We can see the behavior of the parameter $f(d_i^{cloud}, t_i^{fog-cloud})$ from (9) with a Gaussian distribution and standard deviation of 0,15 over the defined mean value. The colors applied for each parameter are the same from the original model. Figure 9(a) shows the energy, throughput and latency for a variable number of incoming packets and fog processing rate. Figure 9(b) exhibits the same parameters for a variable fog-cloud transmission rate. It is possible to note the few changes in the general behavior of the optimized parameters when compared them with Figures 4 and 6. The major difference lies in how the values for a given configuration (see Experimental Design) do not organize into lines, as in Figures 4 and 6. Instead, they are scattered along the surfaces in the solutions space.

*6.3. Maximum Latency Analysis.* The feasibility of a fog node addition is tightly connected to the maximum latency allowed by the specific application. However, an analysis of response time and throughput requirements for a generic UAV application is not possible. This is due to the diverse nature of services and tasks that will impact those requirements. In this sense, a specific scenario is presented to at least indicate how latency requirements can be perceived.

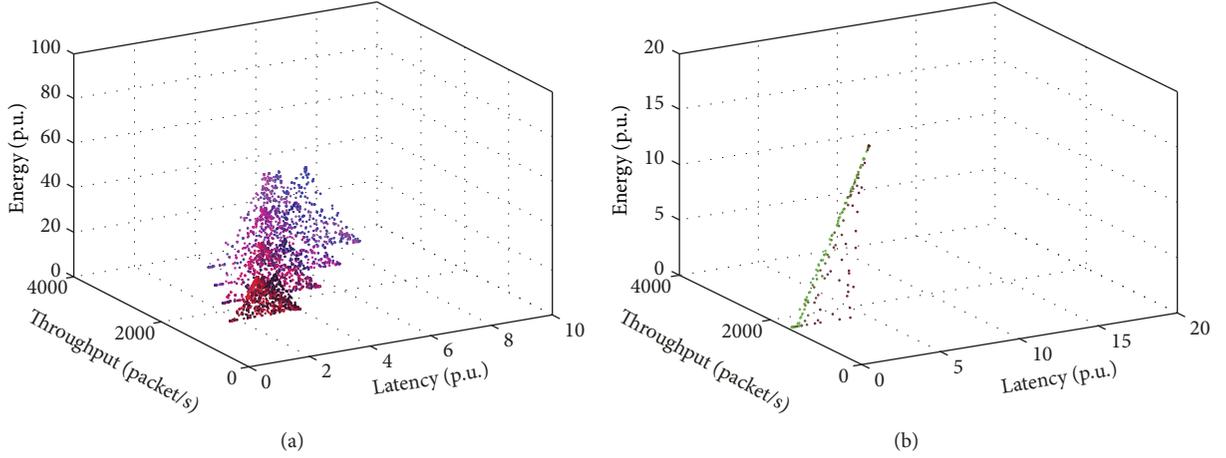

Figure 9: Energy, throughput and latency for (a) a variable number of incoming packets and fog processing rate and (b) a variable fog-cloud transmission rate.

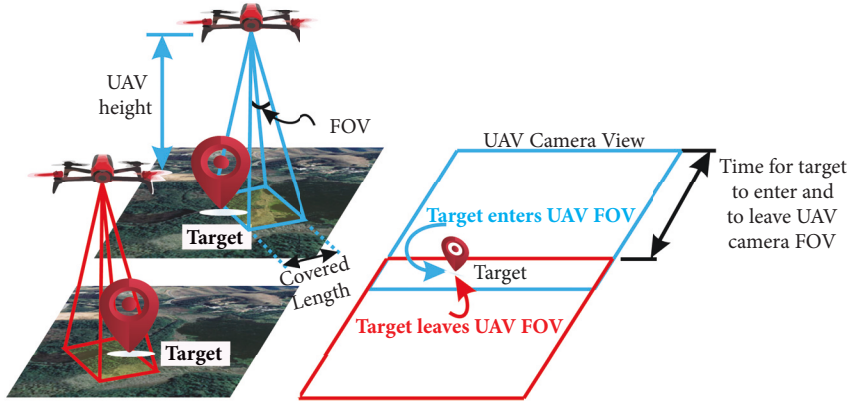

Figure 10: Representation of the target entering and leaving the camera FOV.

For a SAR environment or inspection in a large area, the main goal is to find a specific target. In this way, if the image data is being processed at the cloud, the processing time should not exceed the sum of time that the prominent target enters the camera Field of View (FOV) and leaves it. Thus, in case that the UAV is mapping an area in a specific speed, the decision to track the target should be fast enough to assure that the target is still in the camera FOV during the decision-making. This process is verified in Figure 10.

Considering a camera with 94 degrees of FOV and 3:2 of aspect ratio (e.g., the camera of Phantom 3), then it is possible to determine the length covered at a certain height (h) performed by the camera using [49]

$$lenght = 2 \cdot h \cdot \tan\left(\frac{\tan^{-1}\left((3 \cdot \tan{(FOV)})/2\right)}{2}\right) \quad (10)$$

Then, it is possible to compute the available time for data processing in the established conditions. Figure 11 shows the results for 10, 15 and 20 meters of height considering the total time to capture and to process the data on the UAV as well as to perform the round trip to the network and to process in the fog-cloud computing. The work presented in [50] suggests that a two-hop latency (A2A-A2G) for sensor data can reach 0.84 seconds. Thus, a round trip would have at least 1.68 seconds, which turn some of the UAV cloud applications unfeasible. Despite being strongly correlated with the proposed scenario, these results showed that the deployment of a fog node can benefit the UAV applications in situations where cloud deployment is not feasible.

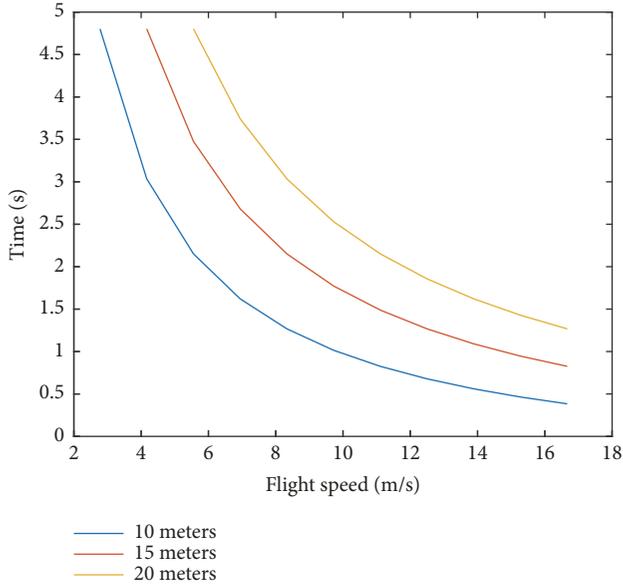

FIGURE 11: Total time between a target entering and leaving the UAV camera FOV for determined flight heights.

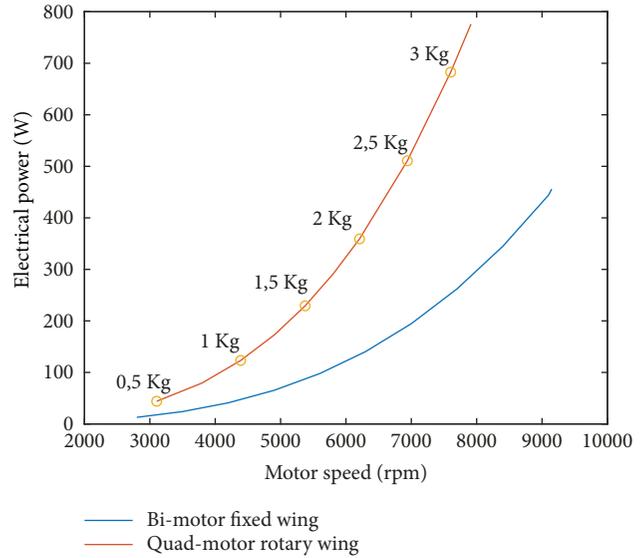

FIGURE 12: Relationship between motor speed and power consumption for rotary and fixed-wing aircraft.

TABLE 5: Electric motor parameters.

| Parameter | Value |
| --- | --- |
| Motor Speed | 1250 rpm/V |
| Current Without Load | 0.6 A |
| Motor Resistance | 0.079 $\Omega$ |
| Maximum Power | 390 W |
| Propeller | Diameter: 254mm Pitch: 119 mm |
| Motor Efficiency | 75 - 85% |

The proposed feasibility model is verified through the comparison between the results of Figure 11 and the ones presented in the previous section. The model should allow the designer to analyze the feasibility of the fog application according to the changes in the model variables. An example would be the analysis of which flight velocity of Figure 10 would make indispensable in a fog computing application.

*6.4. Power Consumption Impact Analysis.* An important analysis in the feasibility evaluation of applying the proposed architecture is the possible impact in energy consumption. Thus, this subsection presents an analysis concerning a study case on the topic. Considering the parameters of Table 5 for an electric motor X2212 1250kV manufacture by Sunnysky [51] and a standard propeller, it is possible to estimate the energy consumption required to lift a quad-motor rotary wing and a bi-motor fixed-wing aircraft from 500 grams up to 3000 grams using the models in [52, 53]. This range of weights was chosen due to the fact that they cover most of the commercial UAVs. Note that the motor was chosen to lift the heaviest aircraft. The same configuration was applied to all weights to simplify the assumptions and comparisons. This does not intend to represent the best design choice or system physical constraints.

Figure 12 presents the relationship between electrical power and the motor speed for quad-motor rotary wing and a bi-motor fixed-wing aircraft. In the quad-motor curve, this figure presents the minimum power required to hover, i.e. to keep the quadrotor flying. This power was computed by the amount of thrust required to overcome gravity for the specific weights shown. The same calculations for the minimum speed/power for a fixed wing are more complex and they largely depend on the aerodynamic aircraft design. As a reference, a 72 dm$^2$ fixed-wing aircraft with drag coefficient of 0.05 at standard temperature and pressure conditions will require between 2400 and 5400 rpm to maintain levelled flight for weight values between 0.5 Kg and 3 Kg. This ultimately results in powers requirements between 13 W and 65 W accordingly to Figure 12.

The energy consumption impact of adding an onboard computer into a UAV is analyzed through this data considering the weight impact from a single board computer and a dedicated battery bank. The power required in the analysis is simplified assuming that the onboard computer has a dedicated isolated battery bank.

For a rotary wing and adding 250 grams, the power required to sustain a flight increases from 9W in a UAV with 0.5 kg to 23,3W in the 3 kg model. For the fixed-wing aircraft, this analysis is more complex and the additional power required to sustain a flight increases around 3 W for the same mentioned design. Note that the added weight significantly impacts the power required to maintain the flight, which consequently affects the energy consumption and aircraft flight time. In this sense, the onboard computer has to be efficient to require lightweight batteries and to reduce the mentioned impact as much as possible. It is also relevant to note that this result is in accordance with other research works in this field. For instance, [54] suggests that the relationship between information processing and communication to the motors consumption is 20/80.

## 7. Conclusions and Future Works

This research work introduced a model to investigate the fog-cloud computing cooperation to overcome the throughput and latency involving multiple aircraft in areas with low communication infrastructure. Besides, the analyzed model includes the power consumption limitations of UAVs in the proposed architecture. This developed architecture assigns a UAV as head coordinator in a fog computing level to process and to control the communications between the nearest nodes and the cloud. Thus, this coordinator manages the computational offloading between cloud and fog. Besides, the head coordinator enables the continuity of the autonomous operation even when it is not connected to the ground station. This process results in a lower latency when dealing with the real-time data provided by the aircraft.

The proposed model differs from previous works by capturing performance behavior not totally explored in UAV scenarios including the limited amount of energy, processing capability restrictions at the fog computing and limited throughput among fog and cloud computing. Those characteristics may also be found in other applications (e.g., IoT) which makes the model even more interesting.

In addition to the detailed model description, an extensive evaluation is reported. The simulations and numerical results showed that the model can be used for fog-cloud computing evaluation when considering latency, energy, and throughput constraints. The model covers three experimental scenarios with an increasing degree of accuracy with respect to the real world.

This work also presented an analysis of scenarios in which cloud computing would not be feasible, indicating the benefits of adding the fog node. The results also indicate that the deployment of a fog node inside a UAV is technically feasible from the energy consumption standpoint. Those results are in accordance with other researchers in this field. Besides, the detailed analysis can be reproduced for other specific study cases.

The architecture layout demonstrated the possibility to overcome the throughput and latency limitations involving multiple aircraft during missions in restricted areas when applying fog-cloud computing cooperation. The contributions of this paper can help researchers understand and design UAVs to further assist missions. Few extensions are foreseen in this research work. First, the architecture will be deployed using real UAVs to further validate the model. Therefore, it is also intended to investigate the impact of other factors such as battery time, number of head coordinators, and type of data processing. Moreover, the comparison with other application models (e.g., IoT) is proposed to show different perspectives for this work.

## Data Availability

The data used to support the findings of this study are included within the article.

## Conflicts of Interest

The authors declare that there are no conflicts of interest regarding the publication of this article.

## Acknowledgments

The authors would like to thank the following Brazilian Federal Agencies: UFJF, CAPES, CNPq, FAPEMIG, INCT-INERGE, and ANEEL/CPFL (PD-02651-0013/2017) for supporting this research work.

## References

[1] D. Giordan, A. Manconi, F. Remondino, and F. Nex, "Use of unmanned aerial vehicles in monitoring application and management of natural hazards," *Geomatics, Natural Hazards and Risk*, vol. 8, no. 1, pp. 1–4, 2017.

[2] M. F. Pinto, A. G. Melo, A. L. M. Marcato, and C. Urdiales, "Case-based reasoning approach applied to surveillance system using an autonomous unmanned aerial vehicle," in *Proceedings of the 26th IEEE International Symposium on Industrial Electronics, ISIE 2017*, pp. 1324–1329, UK, June 2017.

[3] E. Casella, A. Rovere, A. Pedroncini et al., "Drones as tools for monitoring beach topography changes in the Ligurian Sea (NW Mediterranean)," *Geo-Marine Letters*, vol. 36, no. 2, pp. 151–163, 2016.

[4] T. D. Stek, "Drones over Mediterranean landscapes. The potential of small UAV's (drones) for site detection and heritage management in archaeological survey projects: A case study from Le Pianelle in the Tappino Valley, Molise (Italy)," *Journal of Cultural Heritage*, vol. 22, pp. 1066–1071, 2016.

[5] V. Dias, R. Moreira, W. Meira, and D. Guedes, "Diagnosing Performance Bottlenecks in Massive Data Parallel Programs," in *Proceedings of the 16th IEEE/ACM International Symposium on Cluster, Cloud, and Grid Computing, CCGrid 2016*, pp. 273–276, Colombia, May 2016.

[6] R. Mahmud and R. Buyya, Fog computing: A taxonomy survey and future directions, http://arxiv.org/abs/1611.05539.

[7] J. Li, J. Jin, D. Yuan, M. Palaniswami, and K. Moessner, "EHOPES: Data-centered Fog platform for smart living," in *Proceedings of the 25th International Telecommunication Networks and Applications Conference, ITNAC 2015*, pp. 308–313, Australia, November 2015.

[8] A. V. Dastjerdi and R. Buyya, "Fog Computing: Helping the Internet of Things Realize Its Potential," *The Computer Journal*, vol. 49, no. 8, Article ID 7543455, pp. 112–116, 2016.

[9] F. Jalali, K. Hinton, R. Ayre, T. Alpcan, and R. S. Tucker, "Fog computing may help to save energy in cloud computing," *IEEE Journal on Selected Areas in Communications*, vol. 34, no. 5, pp. 1728–1739, 2016.

[10] H. R. Arkian, A. Diyanat, and A. Pourkhalili, "MIST: Fog-based data analytics scheme with cost-efficient resource provisioning for IoT crowdsensing applications," *Journal of Network and Computer Applications*, vol. 82, pp. 152–165, 2017.

[11] L. Liu, Z. Chang, X. Guo, S. Mao, and T. Ristaniemi, "Multiobjective Optimization for Computation Offloading in Fog Computing," *IEEE Internet of Things Journal*, vol. 5, no. 1, pp. 283–294, 2018.

[12] C. C. Byers, "Architectural Imperatives for Fog Computing: Use Cases, Requirements, and Architectural Techniques for Fog-Enabled IoT Networks," *IEEE Communications Magazine*, vol. 55, no. 8, pp. 14–20, 2017.

[13] R. Deng, R. Lu, C. Lai, and T. H. Luan, "Towards power consumption-delay tradeoff by workload allocation in cloud-fog computing," in *Proceedings of the IEEE International Conference on Communications, ICC 2015*, pp. 3909–3914, UK, June 2015.

[14] S. Sarkar, S. Chatterjee, and S. Misra, "Assessment of the Suitability of Fog Computing in the Context of Internet of Things," *IEEE Transactions on Cloud Computing*, vol. 6, no. 1, pp. 46–59, 2015.

[15] S. Sarkar and S. Misra, "Theoretical modelling of fog computing: A green computing paradigm to support IoT applications," *IET Networks*, vol. 5, no. 2, pp. 23–29, 2016.

[16] D. Erdos, A. Erdos, and S. E. Watkins, "An experimental UAV system for search and rescue challenge," *IEEE Aerospace and Electronic Systems Magazine*, vol. 28, no. 5, pp. 32–37, 2013.

[17] J. Gu, T. Su, Q. Wang, X. Du, and M. Guizani, "Multiple Moving Targets Surveillance Based on a Cooperative Network for Multi-UAV," *IEEE Communications Magazine*, vol. 56, no. 4, pp. 82–89, 2018.

[18] B. Schlotfeldt, D. Thakur, N. Atanasov, V. Kumar, and G. J. Pappas, "Anytime Planning for Decentralized Multirobot Active Information Gathering," *IEEE Robotics and Automation Letters*, vol. 3, no. 2, pp. 1025–1032, 2018.

[19] S. W. Loke, *The Internet of Flying-Things: Opportunities and Challenges with Airborne Fog Computing and Mobile Cloud in the Clouds*, 2015, http://arxiv.org/abs/1507.04492.

[20] 3GPP Team, *3GPP: A global initiative*, 2018, http://www.3gpp.org/.

[21] Z. Hao, E. Novak, S. Yi, and Q. Li, "Challenges and Software Architecture for Fog Computing," *IEEE Internet Computing*, vol. 21, no. 2, pp. 44–53, 2017.

[22] Y. Sun, T. Dang, and J. Zhou, "User scheduling and cluster formation in fog computing based radio access networks," in *Proceedings of the 16th IEEE International Conference on Ubiquitous Wireless Broadband, ICUWB 2016*, China, October 2016.

[23] Digi, *Digi XBee® Family Features Comparison Chart*, 2018, https://www.digi.com/pdf/chart_xbee_rf_features.pdf.

[24] J. Chen, J. Xie, Y. Gu et al., "Long-Range and Broadband Aerial Communication Using Directional Antennas (ACDA): Design and Implementation," *IEEE Transactions on Vehicular Technology*, vol. 66, no. 12, pp. 10793–10805, 2017.

[25] M. Malik, F. Farahmand, P. Otto et al., "Architecture Exploration for Energy-Efficient Embedded Vision Applications: From General Purpose Processor to Domain Specific Accelerator," in *Proceedings of the 2016 IEEE Computer Society Annual Symposium on VLSI (ISVLSI)*, pp. 559–564, Pittsburgh, PA, USA, July 2016.

[26] S. S.-D. Xu and T.-C. Chang, "A feasible architecture for ARM-based microserver systems considering energy efficiency," *IEEE Access*, vol. 5, pp. 4611–4620, 2017.

[27] K. Bilal and A. Erbad, "Impact of multiple video representations in live streaming: A cost, bandwidth, and QoE analysis," in *Proceedings of the 2017 IEEE International Conference on Cloud Engineering, IC2E 2017*, pp. 88–94, Canada, April 2017.

[28] P. Houze, E. Mory, G. Texier, and G. Simon, "Applicative-layer multipath for low-latency adaptive live streaming," in *Proceedings of the ICC 2016 - 2016 IEEE International Conference on Communications*, pp. 1–7, Kuala Lumpur, Malaysia, May 2016.

[29] Z. Lin and H. H. Liu, "Topology-based distributed optimization for multi-UAV cooperative wildfire monitoring," *Optimal Control Applications and Methods*, vol. 39, no. 4, pp. 1530–1548, 2018.

[30] J. L. Sanchez-Lopez, M. Molina, H. Bavle, C. Sampedro, R. A. Suárez Fernández, and P. Campoy, "A Multi-Layered Component-Based Approach for the Development of Aerial Robotic Systems: The Aerostack Framework," *Journal of Intelligent & Robotic Systems*, vol. 88, no. 2-4, pp. 683–709, 2017.

[31] C. Luo, J. Nightingale, E. Asemota, and C. Grecos, "A UAV-Cloud System for Disaster Sensing Applications," in *Proceedings of the 2015 IEEE 81st Vehicular Technology Conference (VTC Spring)*, pp. 1–5, Glasgow, United Kingdom, May 2015.

[32] S. Y. M. Mahmoud and N. Mohamed, "Toward a cloud platform for UAV resources and services," in *Proceedings of the 4th IEEE Symposium on Network Cloud Computing and Applications, NCCA 2015*, pp. 23–30, Germany, June 2015.

[33] M. Selecky, M. Rollo, P. Losiewicz, J. Reade, and N. Maida, "Framework for incremental development of complex unmanned aircraft systems," in *Proceedings of the 2015 Integrated Communication, Navigation, and Surveillance Conference (ICNS)*, pp. J3-1–J3-9, Herdon, VA, USA, April 2015.

[34] S. Emel'yanov, D. Makarov, A. I. Panov, and K. Yakovlev, "Multilayer cognitive architecture for UAV control," *Cognitive Systems Research*, vol. 39, pp. 58–72, 2016.

[35] C. Mouradian, D. Naboulsi, S. Yangui, R. H. Glitho, M. J. Morrow, and P. A. Polakos, "A Comprehensive Survey on Fog Computing: State-of-the-art and Research Challenges," *IEEE Communications Surveys & Tutorials*, 2017.

[36] L. M. Vaquero and L. Rodero-Merino, "Finding your way in the fog: Towards a comprehensive definition of fog computing," *Computer Communication Review*, vol. 44, no. 5, pp. 27–32, 2014.

[37] M. Yannuzzi, R. Milito, R. Serral-Gracia, D. Montero, and M. Nemirovsky, "Key ingredients in an IoT recipe: fog computing, cloud computing, and more fog computing," in *Proceedings of the IEEE 19th International Workshop on Computer Aided Modeling and Design of Communication Links and Networks (CAMAD '14)*, pp. 325–329, Athens, Greece, December 2014.

[38] S. Yi, C. Li, and Q. Li, "A survey of fog computing: concepts, applications and issues," in *Proceedings of the Workshop on Mobile Big Data (Mobidata '15)*, pp. 37–42, ACM, Hangzhou, China, June 2015.

[39] P. K. Sharma, M.-Y. Chen, and J. H. Park, "A Software Defined Fog Node Based Distributed Blockchain Cloud Architecture for IoT," *IEEE Access*, vol. 6, pp. 115–124, 2018.

[40] V. Mushunuri, A. Kattepur, H. K. Rath, and A. Simha, "Resource optimization in fog enabled IoT deployments," in *Proceedings of the 2nd International Conference on Fog and Mobile Edge Computing, FMEC 2017*, pp. 6–13, Spain, May 2017.

[41] A. Yousefpour, G. Ishigaki, and J. P. Jue, "Fog Computing: Towards Minimizing Delay in the Internet of Things," in *Proceedings of the 1st IEEE International Conference on Edge Computing, EDGE 2017*, pp. 17–24, USA, June 2017.

[42] L. Gupta, R. Jain, and G. Vaszkun, "Survey of Important Issues in UAV Communication Networks," *IEEE Communications Surveys & Tutorials*, vol. 18, no. 2, pp. 1123–1152, 2016.

[43] S. Mahmoud and N. Mohamed, "Broker architecture for collaborative UAVs cloud computing," in *Proceedings of the 16th*

International Conference on Collaboration Technologies and Systems, CTS 2015*, pp. 212–219, USA, June 2015.

[44] N. Mohamed, J. Al-Jaroodi, I. Jawhar, H. Noura, and S. Mahmoud, "UAVFog: A UAV-based fog computing for Internet of Things," *IEEE SmartWorld, Ubiquitous Intelligence & Computing, Advanced & Trusted Computed, Scalable Computing & Communications, Cloud & Big Data Computing, Internet of People and Smart City Innovation*, pp. 1–8, 2017.

[45] Y. Liu, J. E. Fieldsend, and G. Min, "A Framework of Fog Computing: Architecture, Challenges, and Optimization," *IEEE Access*, vol. 5, pp. 25445–25454, 2017.

[46] T. Wang, J. Zeng, Y. Lai et al., "Data collection from WSNs to the cloud based on mobile Fog elements," *Future Generation Computer Systems*, 2017.

[47] S. Kitanov and T. Janevski, "Fog Computing as a Support for 5G Network," *Journal of Emerging research and solutions in ICT*, vol. 1, no. 2, pp. 47–59, 2016.

[48] M. Works, *Matlab User Manual Version R2016b*, Math Works: Natick, MA, USA, 2016.

[49] C. Yu, J. Wang, J. Shan, and M. Xin, "Multi-UAV UWA video surveillance system," in *Proceedings of the 14th International Conference on Control, Automation, Robotics and Vision, ICARCV 2016*, Thailand, November 2016.

[50] Y. Zhou, N. Cheng, N. Lu, and X. S. Shen, "Multi-UAV-Aided Networks: Aerial-Ground Cooperative Vehicular Networking Architecture," *IEEE Vehicular Technology Magazine*, vol. 10, no. 4, pp. 36–44, 2015.

[51] SunnySky, *X2212 motor datasheet*, China, 2018, http://www.rcsunnysky.com.

[52] F. Morbidi, R. Cano, and D. Lara, "Minimum-energy path generation for a quadrotor UAV," in *Proceedings of the 2016 IEEE International Conference on Robotics and Automation*, pp. 1492–1498, Stockholm, Sweden, March 2016.

[53] D. H. Choi, S. H. Kim, and D. K. Sung, "Energy-efficient maneuvering and communication of a single UAV-based relay," *IEEE Transactions on Aerospace and Electronic Systems*, vol. 50, no. 3, pp. 2320–2327, 2014.

[54] A. Trotta, M. D. Felice, F. Montori, K. R. Chowdhury, and L. Bononi, "Joint Coverage, Connectivity, and Charging Strategies for Distributed UAV Networks," *IEEE Transactions on Robotics*, vol. 34, no. 4, pp. 883–900, 2018.

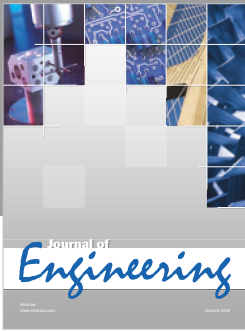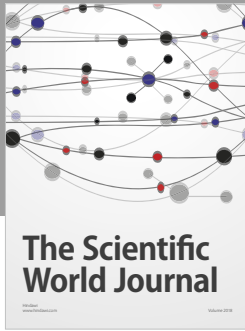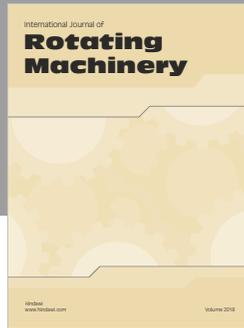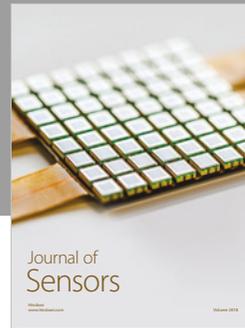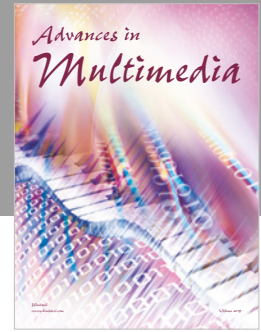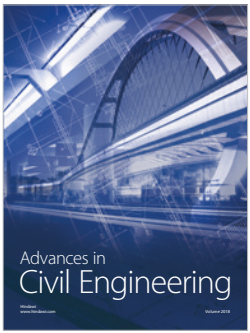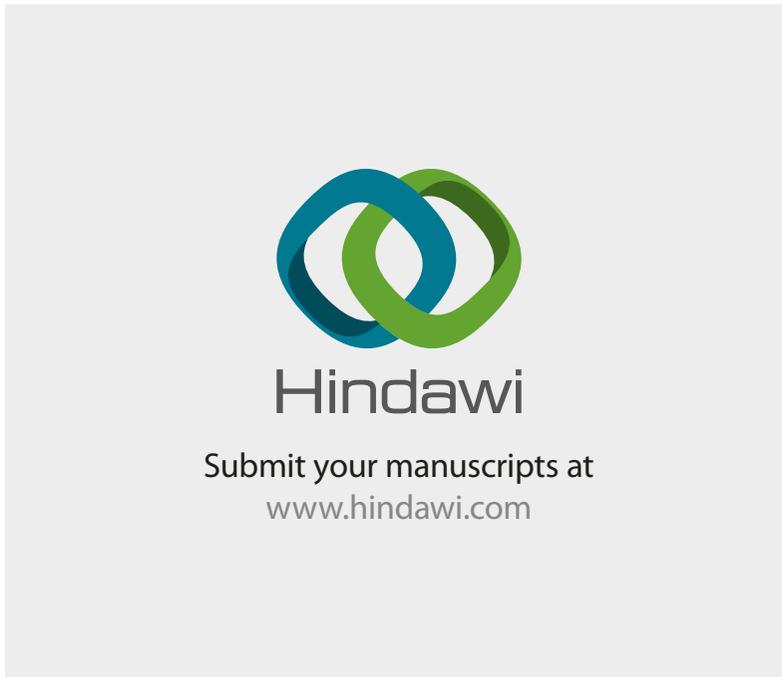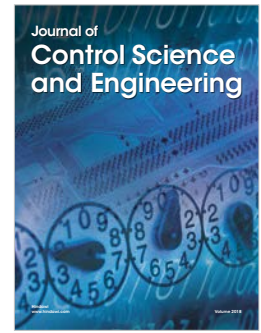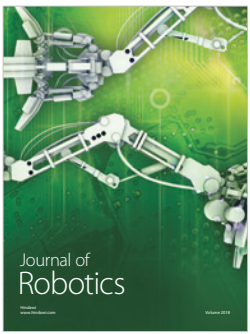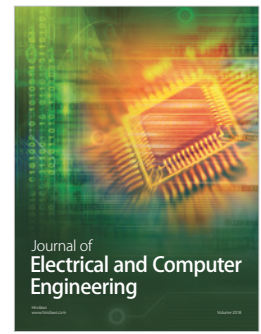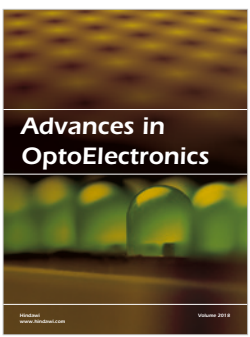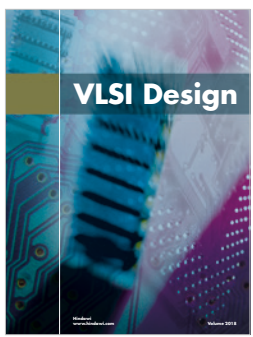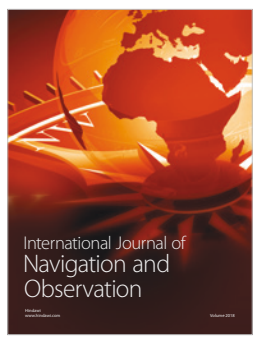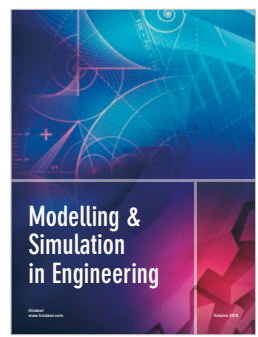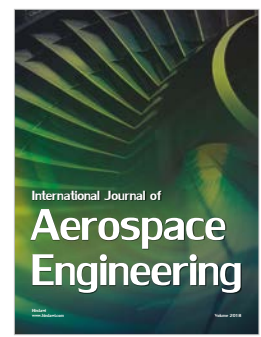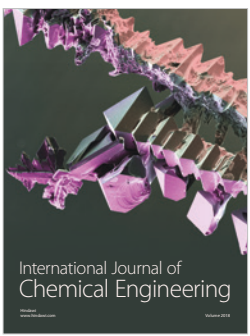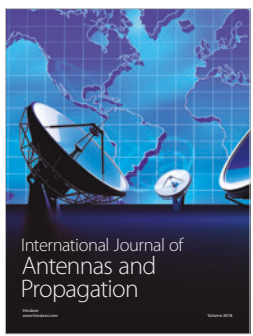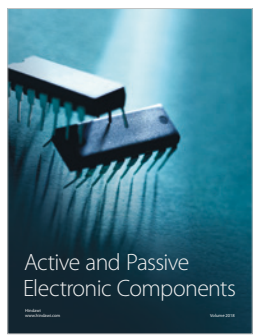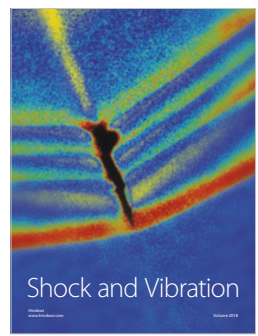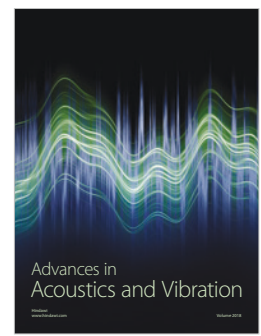